%% file: main.tex
\documentclass[10pt,twocolumn,letterpaper]{article}

\usepackage[preprint]{cvpr}      

\input{preamble}

\newcommand{\method}{VerdictAD\xspace}
\usepackage[table]{xcolor}
\usepackage{multirow}
\definecolor{verdictblue}{RGB}{232,243,252}

\definecolor{cvprblue}{rgb}{0.21,0.49,0.74}
\usepackage[pagebackref,breaklinks,colorlinks,allcolors=cvprblue]{hyperref}

\def\confName{CVPR}
\def\confYear{2027}

\title{VerdictAD: Few-Shot Anomaly Detection via Decision-Guided Refinement}

\author{
Shangquan Sun$^{1}$ \quad
Wenjun Luo$^{1}$ \quad
Yuanzhi Wang$^{1}$ \quad
Ruiqi Li$^{1}$ \\
Ruoyu Chen$^{2}$ \quad
Jiachen Liu$^{3}$ \quad
Shijian Lu$^{1}$ \\
$^{1}$Nanyang Technological University \quad
$^{2}$Michigan State University \quad
$^{3}$University of Michigan
}

\begin{document}
\maketitle

\begin{abstract}
Detecting defects in real-world industrial inspection often requires learning from only a few normal images per category.
Recent few-shot anomaly detection methods have achieved strong performance by measuring feature differences between query images and normal references.
However, visual differences do not always indicate genuine defects, as normal appearance variations can also produce high anomaly scores.
Conversely, structural or logical anomalies may not produce strong local feature differences, making them difficult to detect.
To address this limitation, we introduce \method, a training-free framework that refines existing anomaly detectors through visual decisions.
Rather than replacing existing detectors, \method adopts a typed decision model to assess the query image and its local regions against a few normal references.
By directly producing structured decision probabilities without text generation, the model provides complementary evidence for refining anomaly responses at different spatial scales.
Building on this capability, we introduce two modules: Global Verdict Refinement (GVR) uses image-level decisions to refine the overall anomaly score, while Local Verdict Refinement (LVR) aggregates regional decisions to refine the anomaly map.
Combined with a simple DINOv2-based detector, \method achieves competitive performance in one-shot and few-shot anomaly detection.
More importantly, as a plug-and-play framework, it consistently improves existing anomaly detectors across MVTec AD, VisA, and MVTec LOCO AD, achieving strong results across one-shot and few-shot settings.
Our code and experimental logs are publicly available at
\url{https://github.com/sgstq0811/VerdictAD-Jev-AD}.
\end{abstract} 
\input{sec/1_intro}

\input{sec/2_related}

\input{sec/3_method}

\input{sec/4_experiment}

\input{sec/5_result}

\input{sec/6_conclu}
\input{sec/X_LLM_usage}
{
    \small
    \bibliographystyle{ieeenat_fullname}
    \bibliography{main}
}


\end{document}

%% file: preamble.tex
\usepackage[most]{tcolorbox}
\definecolor{promptbg}{rgb}{0.95,0.96,0.98}
\definecolor{promptline}{rgb}{0.21,0.49,0.74}
\newtcolorbox{promptbox}[1]{enhanced, breakable, colback=promptbg, colframe=promptline, boxrule=0.5pt, arc=2pt,
  left=4pt, right=4pt, top=3pt, bottom=3pt, fonttitle=\small\bfseries, title={#1}, fontupper=\small}

%% file: sec/1_intro.tex
\section{Introduction}
Visual anomaly detection (AD) has long been a challenging task in computer vision, particularly in industrial inspection. 
In real-world industry, even minor defects, such as scratches, contamination, or missing components, can affect product quality and cause failures in subsequent manufacturing processes. 
Therefore, automated systems capable of accurately detecting such defects are essential for reliable and cost-effective industrial production.

\input{sec/fig_sec/teaser}

A major challenge in industrial anomaly detection is the limited availability of training data. 
Full-shot methods typically require hundreds of defect-free images per category to model normal appearances, which can be difficult to obtain in practice. 
In contrast, zero-shot methods leverage vision-language models (VLMs) and textual prompts to detect anomalies without normal training samples. 
However, they often struggle with subtle, non-semantic defects, such as tiny cracks, that are difficult to describe through language alone.
In this work, we focus on the more practical yet challenging few-shot setting. 
Under this setting, only a handful of normal images are available for each target category to characterize its normal appearance.

Existing few-shot industrial anomaly detection methods mainly include reconstruction-based, memory-bank-based, and vision-language model (VLM)-based approaches.
Despite their different designs, these methods typically produce image-level anomaly scores and pixel-level anomaly maps to identify and localize defects.
However, visual differences do not always indicate genuine defects, as normal appearance variations can also produce high anomaly scores.
As shown in Fig.~\ref{fig:intro_examples}, a normal juice bottle may be flagged as anomalous because its liquid color differs from the references, while variations in screw orientation and background textures can also trigger false responses.
Conversely, logical anomalies may produce misleading anomaly responses.
In the breakfast box example, missing dried fruits cause a logical anomaly, but the baseline incorrectly highlights the entire tray rather than the actual defective region.
These examples show that anomaly responses based on visual differences do not always accurately reflect genuine defects.

This motivates us to verify anomaly responses rather than solely improving visual representations.
Recent typed decision models~\cite{typesafe2026jev,chen2026introducing} provide probabilities over predefined answers without generating free-form text.
Such decisions offer complementary information to the visual differences measured by existing anomaly detectors.
We therefore explore their use in refining image-level anomaly scores and pixel-level anomaly maps.
To this end, we propose \method, a training-free framework for decision-guided anomaly refinement.
We first build a simple anomaly detector using frozen DINOv2-L features~\cite{oquab2023dinov2} to obtain image-level anomaly scores and pixel-level anomaly maps.
We then introduce two complementary refinement modules.
Global Verdict Refinement (GVR) combines reference-free and reference-conditioned decisions to refine image-level anomaly scores.
Local Verdict Refinement (LVR) evaluates image regions through cropped views and boxed full-image views, producing a verification map to reweight the original anomaly map.
By suppressing misleading responses more strongly than defect-related ones, LVR improves the relative prominence of genuine defects without generating new anomaly evidence.
Together, these modules achieve state-of-the-art performance across multiple few-shot anomaly detection benchmarks.
Moreover, the underlying detector can be replaced with existing anomaly detection methods, allowing \method\ to refine their predictions without modifying their architectures or requiring additional training.
Our main contributions are threefold:
\begin{enumerate}
    \item We propose \method, a training-free framework that leverages typed visual decisions to refine anomaly detection and localization.
    
    \item We introduce GVR and LVR to refine image-level anomaly scores and pixel-level anomaly maps through global and local visual decisions.
    
    \item Extensive experiments on MVTec AD, VisA, and MVTec LOCO AD demonstrate state-of-the-art performance across multiple few-shot settings.
    
    \item We demonstrate the generality of \method\ by consistently improving existing anomaly detectors without additional training.
\end{enumerate}

\input{sec/fig_sec/intro_example}

%% file: sec/fig_sec/teaser.tex
\begin{figure}[t]
    \centering

    \includegraphics[width=\linewidth]{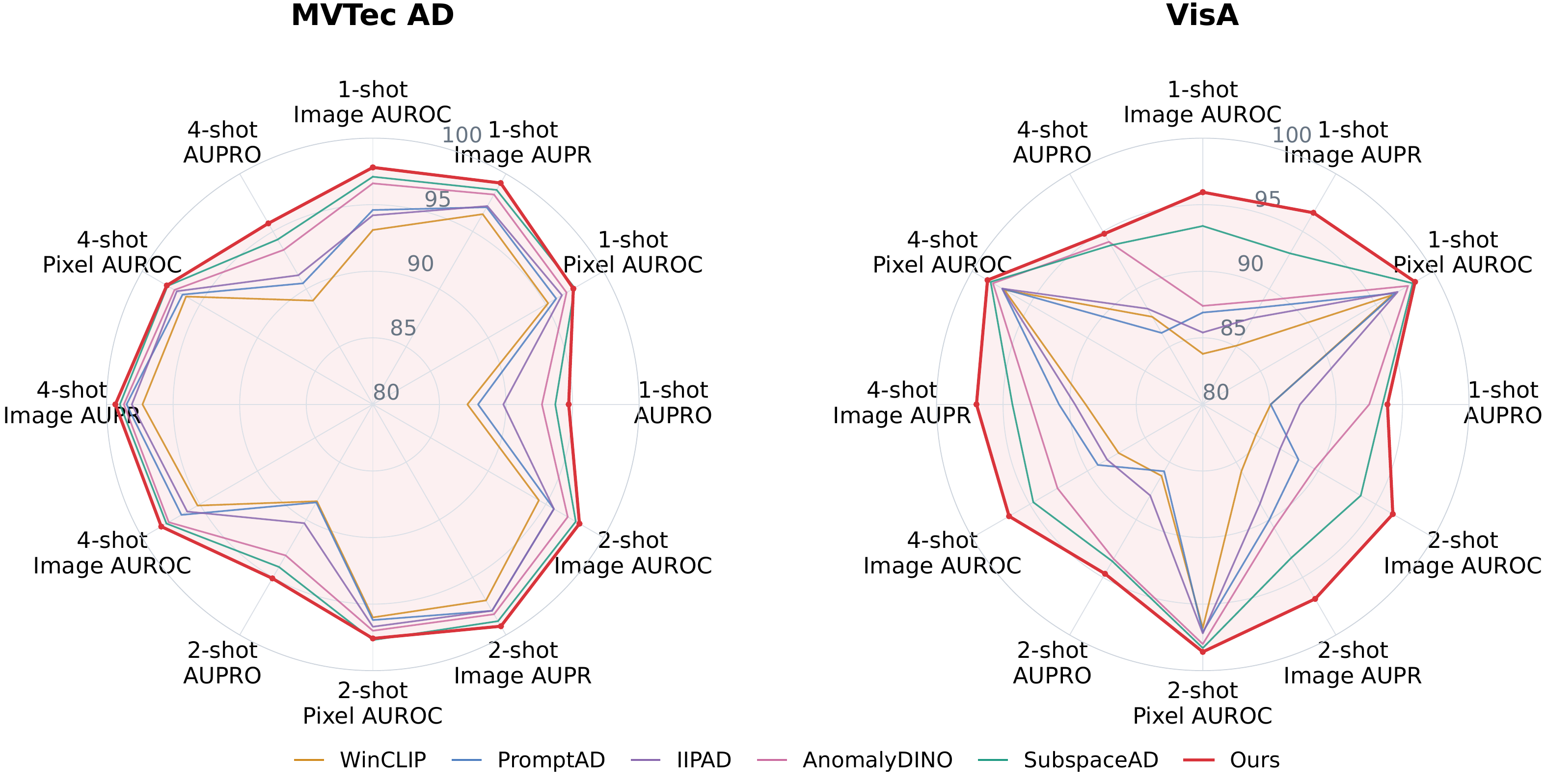}
    \centerline{(a) Few-shot performance}

    \vspace{3mm}

    \includegraphics[width=\linewidth]{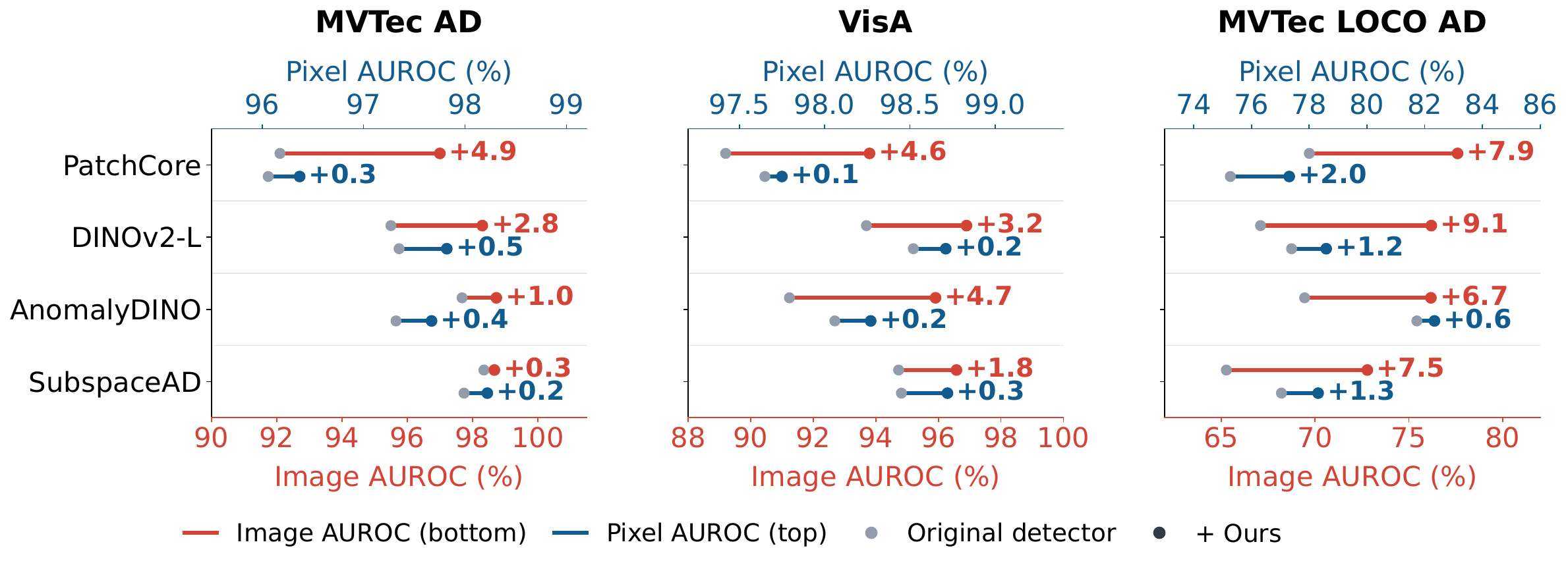}
    \centerline{(b) Improvements across detectors}

    \vspace{-1mm}
    \caption{
    Performance improvements with \method.
    (a) Anomaly detection and localization performance under the 1-, 2-, and 4-shot settings on MVTec AD and VisA.
    (b) Image-level and pixel-level AUROC before and after applying \method\ to different anomaly detectors.
    }
    \label{fig:teaser}
    \vspace{-3mm}
\end{figure}

%% file: sec/fig_sec/intro_example.tex
\begin{figure}[t]
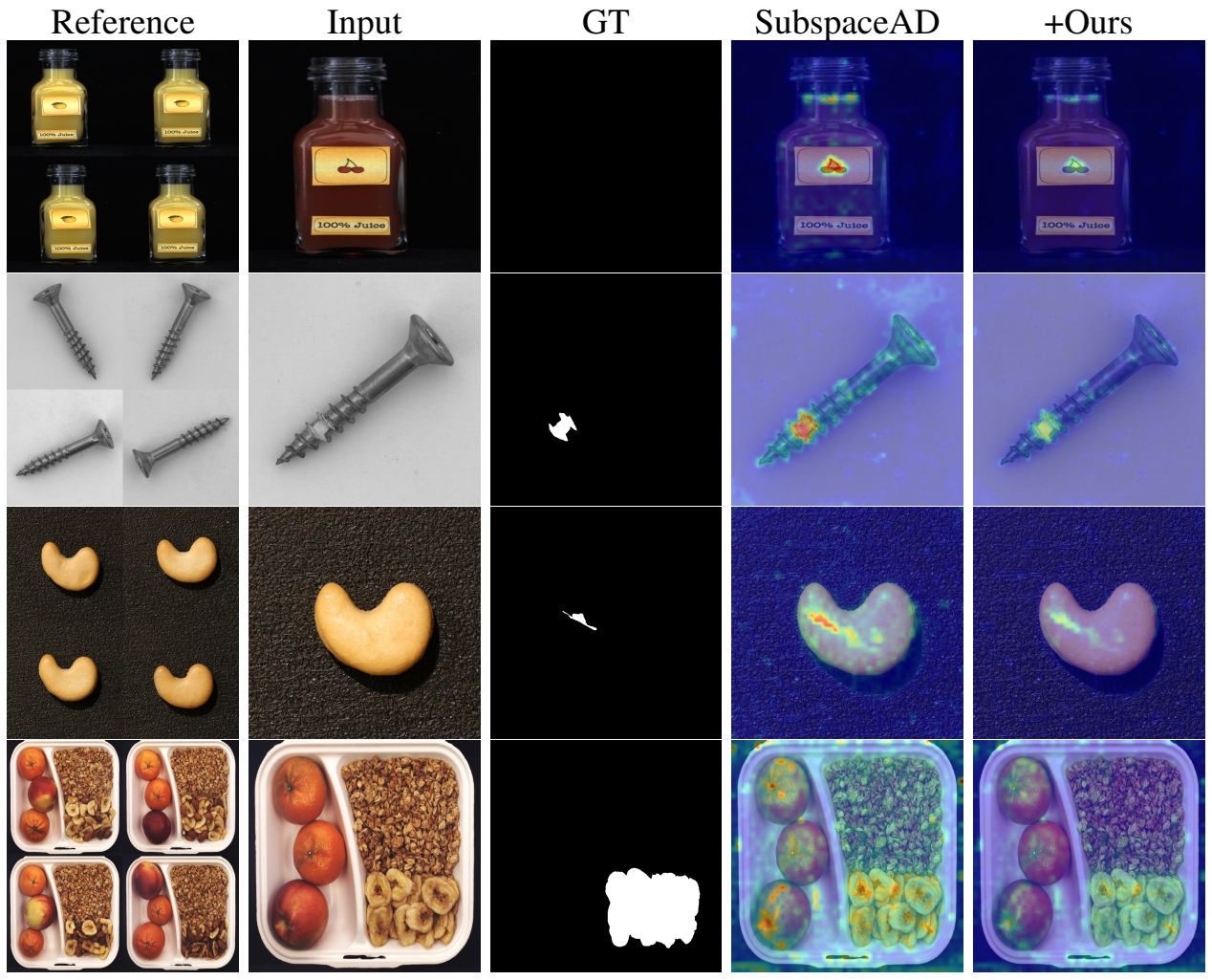

    \centering
    \begingroup
    \setlength{\tabcolsep}{1pt}
    \renewcommand{\arraystretch}{1.0}

    \newcommand{\figroot}{sec/fig_sec/result_fig}
    \newcommand{\caseimg}[2]{%
        \includegraphics[width=0.192\columnwidth]{%
            \figroot/#1/#2.jpg}%
    }

    \scriptsize
    \begin{tabular}{@{}ccccc@{}}
        Reference &
        Input &
        GT &
        SubspaceAD &
        +Ours \\[-1pt]

        \caseimg{loco_juice_bottle_juice_bottle_good_001}{ref} &
        \caseimg{loco_juice_bottle_juice_bottle_good_001}{input} &
        \caseimg{loco_juice_bottle_juice_bottle_good_001}{gt} &
        \caseimg{loco_juice_bottle_juice_bottle_good_001}{baseline_overlay} &
        \caseimg{loco_juice_bottle_juice_bottle_good_001}{ours_overlay}
        \\[-2pt]

        \caseimg{mvtec_screw_006}{ref} &
        \caseimg{mvtec_screw_006}{input} &
        \caseimg{mvtec_screw_006}{gt} &
        \caseimg{mvtec_screw_006}{baseline_overlay} &
        \caseimg{mvtec_screw_006}{ours_overlay}
        \\[-2pt]

        \caseimg{visa_cashew_090}{ref} &
        \caseimg{visa_cashew_090}{input} &
        \caseimg{visa_cashew_090}{gt} &
        \caseimg{visa_cashew_090}{baseline_overlay} &
        \caseimg{visa_cashew_090}{ours_overlay}
        \\[-2pt]
        
        \caseimg{loco_breakfast_box_breakfast_box_logical_anomalies_013}{ref} &
        \caseimg{loco_breakfast_box_breakfast_box_logical_anomalies_013}{input} &
        \caseimg{loco_breakfast_box_breakfast_box_logical_anomalies_013}{gt} &
        \caseimg{loco_breakfast_box_breakfast_box_logical_anomalies_013}{baseline_overlay} &
        \caseimg{loco_breakfast_box_breakfast_box_logical_anomalies_013}{ours_overlay}
        \\
    \end{tabular}
    \endgroup

\vspace{-2mm}
\caption{
Examples of visual differences that produce misleading anomaly responses.
(a) \textbf{Juice bottle:} Different liquid colors lead to false anomaly responses on the bottle logo.
(b) \textbf{Screw:} Variations in screw orientation cause widespread false responses.
(c) \textbf{Cashew:} Textured backgrounds are incorrectly identified as anomalous regions.
(d) \textbf{Breakfast box:} Missing dried fruits cause a logical anomaly, but the baseline incorrectly highlights the entire tray.
\method\ uses visual decisions to suppress misleading responses while preserving useful anomaly evidence.
}
    \label{fig:intro_examples}
    \vspace{-3mm}
\end{figure}

%% file: sec/2_related.tex
\section{Related Work}
\label{sec:related}

\subsection{Industrial Visual Anomaly Detection}
Industrial anomaly detection identifies deviations from normal appearance, typically without annotated defects~\cite{chandola2009anomaly,pang2021deep}.
Early reconstruction approaches use autoencoders to expose defects through reconstruction errors~\cite{bergmann2018improving,zavrtanik2021draem}, while reverse distillation~\cite{deng2022anomaly} and contrastive reconstruction~\cite{guo2023recontrast} operate in learned feature spaces.
Recent reconstruction-based methods employ Transformer architectures~\cite{you2022unified,zhang2023exploring,guo2025dinomaly}, diffusion models~\cite{zhang2025diffusionad,fuvcka2024transfusion,he2024diffusion}, and state-space models~\cite{he2024mambaad}, while prototype-guided reconstruction further improves unified anomaly detection~\cite{luo2025inp}.
Alternatively, flow-based methods estimate feature distributions~\cite{yu2021fastflow,gudovskiy2022cflow}.
Feature matching~\cite{cohen2020sub,roth2022towards} and statistical modeling~\cite{defard2021padim} have also been adopted for anomaly detection, with subsequent work addressing noisy references, speed, and discriminative scoring~\cite{jiang2022softpatch,batzner2024efficientad,liu2023simplenet,zou2022spot,bergman2020deep}.
These approaches are commonly evaluated on industrial benchmarks with varied defects and acquisition conditions~\cite{bergmann2019mvtec,heckler2026mvtec,wang2024real}.

\subsection{\hspace{-0.2mm}Few-shot\hspace{-0.2mm} and\hspace{-0.2mm} Training-free\hspace{-0.2mm} Anomaly\hspace{-0.2mm} Detection\hspace{-0.2mm}}
Few-shot detection seeks to characterize normality from only a handful of examples.
Early adaptations of nearest-neighbor matching and feature distributions include optimized PatchCore, GraphCore, and personalized anomaly models~\cite{santos2023optimizing,xie2023pushing,kwak2024few}.
Strong visual representations from self-supervised transformers and masked image modeling~\cite{caron2021emerging,he2022masked,oquab2023dinov2,simeoni2025dinov3} have enabled increasingly simple detectors.
AnomalyDINO~\cite{damm2025anomalydino} uses patch-level DINOv2 features, FastRecon~\cite{fang2023fastrecon} reconstructs normal features, and SubspaceAD~\cite{lendering2026subspacead} fits a normal feature subspace.
Recent variants investigate prototype~\cite{li2026fastref} and structural feature~\cite{jiang2026defect} refinement, object-centric logical anomalies~\cite{fuvcka2026objectcore}, multi-scale features~\cite{ali2026radiocore}, geometric priors~\cite{xu2026anomalies}, multiview consistency~\cite{nan2026cross}, and deployment robustness~\cite{shinde2026subtledc}.
Related work also seeks transferable detection across categories and domains~\cite{strater2024generalad}.
Unlike methods that modify normal feature representations, our approach uses visual decisions to refine the anomaly scores and maps.

\subsection{Vision-Language Anomaly Detection}
Vision-language representations such as CLIP~\cite{radford2021learning} support anomaly detection through text-image similarity.
WinCLIP~\cite{jeong2023winclip} introduces window-level prompting, while PromptAD~\cite{li2024promptad}, instance-induced prompting~\cite{lv2025one}, and multimodal prompt learning~\cite{lee2026bidirectional} adapt the prompt space to few-shot normal references.
Zero-shot approaches leverage prompt learning~\cite{zhou2024anomalyclip,cao2024adaclip,wang2026code}, vision-language adaptation~\cite{deng2023anovl,xu2024customizing,chen2023zero}, and mutual scoring across unlabeled images~\cite{li2024musc}.
Recent work also adapts vision foundation models for zero-shot anomaly detection~\cite{fuvcka2026anomalyvfm}.
Recent work further introduces image-level-supervised reasoning~\cite{jin2026reasoning}, knowledge-grounded reasoning~\cite{zhang2026adseeker}, and agentic vision-language models~\cite{belhadi2026vla}.
However, these approaches typically use language-aligned representations or reasoning to construct an anomaly detector.
In contrast, \method uses a frozen typed decision model to produce direct anomaly judgments that refine image-level scores and pixel-level maps from existing detectors, without additional training or changes to their architectures.

\subsection{Typed Decision Models}
Typed decision models have recently emerged as an alternative to generative language models for structured decision-making, building on earlier work in verification and LLM-as-a-Judge~\cite{cobbe2021training,zheng2023judging}.
Unlike conventional LLMs that produce free-form text, Jev~\cite{typesafe2026jev} directly answers typed questions with probabilities over predefined options.
Recent open models, including NeoHorse-Jev~\cite{tokenrhythm2026neohorsejev,team2026neohorse} and JevRL~\cite{tegong2026jevrl}, further explore decision modeling and learning.
Recent open-source decision models built upon multimodal backbones have further extended typed decision-making to visual inputs~\cite{chen2026introducing,cloudflare2026clefflash,qwen2026qwen35}.
Recent studies have further explored their applications in alignment assessment~\cite{guo2026just}, adversarial robustness evaluation~\cite{hu2026jevadvbench}, and reinforcement learning~\cite{ma2026can}.
Despite these advances, typed decision models remain largely unexplored for industrial anomaly detection.
Unlike methods that directly predict anomalies, \method leverages typed decisions as complementary evidence to refine image-level scores and pixel-level anomaly maps from existing detectors.

%% file: sec/3_method.tex
\section{Method}
\label{sec:method}


\input{sec/fig_sec/pipeline}

\subsection{Preliminaries}
\label{sec:prelim}

\vspace{1mm}
\noindent\textbf{Few-shot Anomaly Detection.}
Given a query image $\mathbf{X}\in\mathbb{R}^{H\times W\times 3}$ and $k$ normal reference images $\mathcal{R}=\{\mathbf{R}_j\}_{j=1}^{k}$, where $k\in\{1,2,4\}$, a base anomaly detector $\mathcal{D}$ produces an image-level anomaly score $s$ and a pixel-level anomaly map $\mathbf{M}\in\mathbb{R}_{\ge 0}^{H\times W}$:
\begin{equation}
  (s, \mathbf{M}) = \mathcal{D}(\mathbf{X};\mathcal{R}).
  \label{eq:base}
\end{equation}
The score $s$ is typically obtained by pooling the anomaly map, such as max pooling in PatchCore~\cite{santos2023optimizing} and top-$1\%$ averaging in AnomalyDINO~\cite{damm2025anomalydino}.
Despite these differences, \method serves as a plug-and-play refinement framework that operates on both outputs without modifying the underlying detector.
Our refinement framework operates on both outputs without modifying the base detector.

\vspace{1mm}
\noindent\textbf{Typed Decision Models.}
Unlike conventional generative language models, typed decision models such as Jev~\cite{typesafe2026jev,chen2026introducing} produce probabilities over predefined answers without text generation.
A model $\mathcal{J}$ takes a set of images $\mathcal{I}$ and a question schema $\mathcal{Q}$ as input.
Among the supported Choice, Score, and Noul modes, we use Noul to obtain binary anomaly decisions.
Specifically, we adopt the frozen Clef-Flash model~\cite{cloudflare2026clefflash} to obtain decision logits without autoregressive text generation.

\subsection{Overall Pipeline}
\label{sec:pipeline}

As illustrated in Fig.~\ref{fig:pipeline}, \method consists of a base anomaly detector and two decision-guided refinement modules, GVR and LVR.

\vspace{1mm}
\noindent\textbf{Base Anomaly Detector.}
We construct a simple anomaly detector using a pretrained DINOv2-L encoder~\cite{oquab2023dinov2}.
Given a query image and $k$ normal reference images, we extract patch features from an intermediate transformer layer of the frozen encoder.
For each query patch, we compute its nearest-neighbor distance to the reference patch features to construct an anomaly map $\mathbf{M}$.
The image-level anomaly score $s$ is obtained by max pooling over $\mathbf{M}$.

\vspace{1mm}
\noindent\textbf{Decision-Guided Refinement.}
Given the initial anomaly score and map, GVR uses a frozen typed decision model to evaluate the query image with and without normal references, producing a decision probability to refine the image-level score.
LVR further evaluates local regions through two complementary views: cropped regions that emphasize local details and boxed full-image views that preserve global context.
The resulting regional decisions are aggregated into a verification map to reweight the initial anomaly map.
Since both modules operate on the detector outputs, the base detector can be replaced with existing anomaly detection methods without additional training.

\subsection{Global Verdict Refinement}
\label{sec:GVR}

GVR combines reference-free and reference-conditioned decisions to refine the image-level anomaly score:
\begin{align}
  z_{\mathrm{free}} &= \mathcal{J}(\{\mathbf{X}\}, q_{\mathrm{free}}),
  \label{eq:zfree}\\
  z_{\mathrm{ref}} &= \mathcal{J}\big(\{\mathbf{X}, \operatorname{Tile}(\mathcal{R})\}, q_{\mathrm{ref}}\big),
  \label{eq:zref}\\
  z &= \tfrac{1}{2}(z_{\mathrm{free}} + z_{\mathrm{ref}}),
  \label{eq:zmean}
\end{align}
where $\operatorname{Tile}(\cdot)$ arranges the normal references into a single grid, and $q_{\mathrm{free}}$ and $q_{\mathrm{ref}}$ denote the corresponding binary decision questions.
The reference-free logit $z_{\mathrm{free}}$ captures general defect evidence, while the reference-conditioned logit $z_{\mathrm{ref}}$ assesses anomalies relative to the normal references.
We use their averaged logit $z$ to refine the detector score:
\begin{equation}
  S = s \cdot \sigma(z)^{\alpha},
  \label{eq:gate}
\end{equation}
where $\sigma(\cdot)$ denotes the sigmoid function and $\alpha$ controls the strength of image-level refinement.
The decision probability acts as a multiplicative gate, suppressing anomaly scores that receive weak support from the decision model.
This allows normal appearance variations to be downweighted while preserving stronger responses to genuine defects.
When $\alpha=0$, the original detector score is recovered
For zero-shot detectors without references ($\mathcal{R}=\varnothing$), we use $z=z_{\mathrm{free}}$.

\subsection{Local Verdict Refinement}
\label{sec:LVR}

While GVR refines image-level anomaly scores, LVR evaluates anomaly responses at the regional level.
We place $N$ overlapping square windows with a predefined window size and stride over the query image.
Each window is represented by $b_i=(x_i^0,y_i^0,x_i^1,y_i^1)\in[0,1]^4$, where $i=1,\dots,N$, and $\Omega(b_i)$ denotes its pixel region.
We denote the corresponding image crop by $\mathbf{X}[b_i]$.
Each window is evaluated using two complementary views.

\vspace{1mm}
\noindent\textbf{Crop Decision.}
We compare the query crop $\mathbf{X}[b_i]$ with the corresponding reference crops $\mathcal{R}[b_i]=\{\mathbf{R}_j[b_i]\}_{j=1}^{k}$:
\begin{equation}
  z^{\mathrm{c}}_i = \mathcal{J}\big(\{\mathbf{X}[b_i], \operatorname{Tile}(\mathcal{R}[b_i])\}, q_{\mathrm{crop}}\big),
  \label{eq:zcrop}
\end{equation}
where $q_{\mathrm{crop}}$ is the binary question for crop-based anomaly assessment.
Cropping emphasizes local details, making it suitable for defects such as scratches, cracks, and holes.

\input{sec/fig_sec/qualitative_comparison}

\vspace{1mm}
\noindent\textbf{Box Decision.}
We draw a red box around $b_i$ on the full query image, denoted by $\operatorname{Box}(\mathbf{X},b_i)$, and compare it with the full references:
\begin{equation}
  z^{\mathrm{b}}_i = \mathcal{J}\big(\{\operatorname{Box}(\mathbf{X},b_i), \operatorname{Tile}(\mathcal{R})\}, q_{\mathrm{box}}\big),
  \label{eq:zbox}
\end{equation}
where $q_{\mathrm{box}}$ is the corresponding binary decision question.
Unlike cropping, boxing preserves global context, making it suitable for logical anomalies involving missing, extra, or misplaced components.
We use red boxes throughout our experiments, with the effect of box color examined in supplementary materials.

\vspace{1mm}
\noindent\textbf{Verification Map Aggregation.}
We combine the crop and box decisions into a set of $2N$ regional logits:
\begin{equation}
  z_i \!=\!
  \begin{cases}
    z_i^{\mathrm{c}}, & 1\!\le\! i\!\le\! N,\\
    z_{i-N}^{\mathrm{b}}, & N\!<\!i\!\le\! 2N,
  \end{cases}
  \quad b_{i+N}\!=\!b_i \ (1\!\le\! i\!\le\! N).
  \label{eq:merge}
\end{equation}
The crop and box decisions for each window share the same spatial coordinates.
We construct a verification map by averaging the transformed decision probabilities of all windows covering each pixel $u$:
\begin{equation}
  \mathbf{W}(u) =
  \frac{\sum_{i=1}^{2N}\mathbb{I}[u\in\Omega(b_i)]\,\sigma(z_i)^{\gamma}}
  {\sum_{i=1}^{2N}\mathbb{I}[u\in\Omega(b_i)]},
  \label{eq:W}
\end{equation}
where $\mathbb{I}[\cdot]$ is the indicator function and $\gamma$ controls the strength of pixel-level refinement.
The original anomaly map is then reweighted by the verification map:
\begin{equation}
  \mathbf{M}'(u) = \mathbf{M}(u)\cdot\mathbf{W}(u).
  \label{eq:write}
\end{equation}
Since $\mathbf{W}(u)\in(0,1)$, LVR suppresses responses in regions judged less likely to contain defects while retaining relatively stronger responses in anomalous regions.
When $\gamma=0$, the original anomaly map is recovered.
As the refinement is multiplicative, $\mathbf{M}'(u)=0$ wherever $\mathbf{M}(u)=0$, meaning that LVR can only reweight existing anomaly evidence rather than introduce new responses.

%% file: sec/fig_sec/pipeline.tex
\begin{figure*}[t]
  \centering
  \includegraphics[width=1\textwidth]{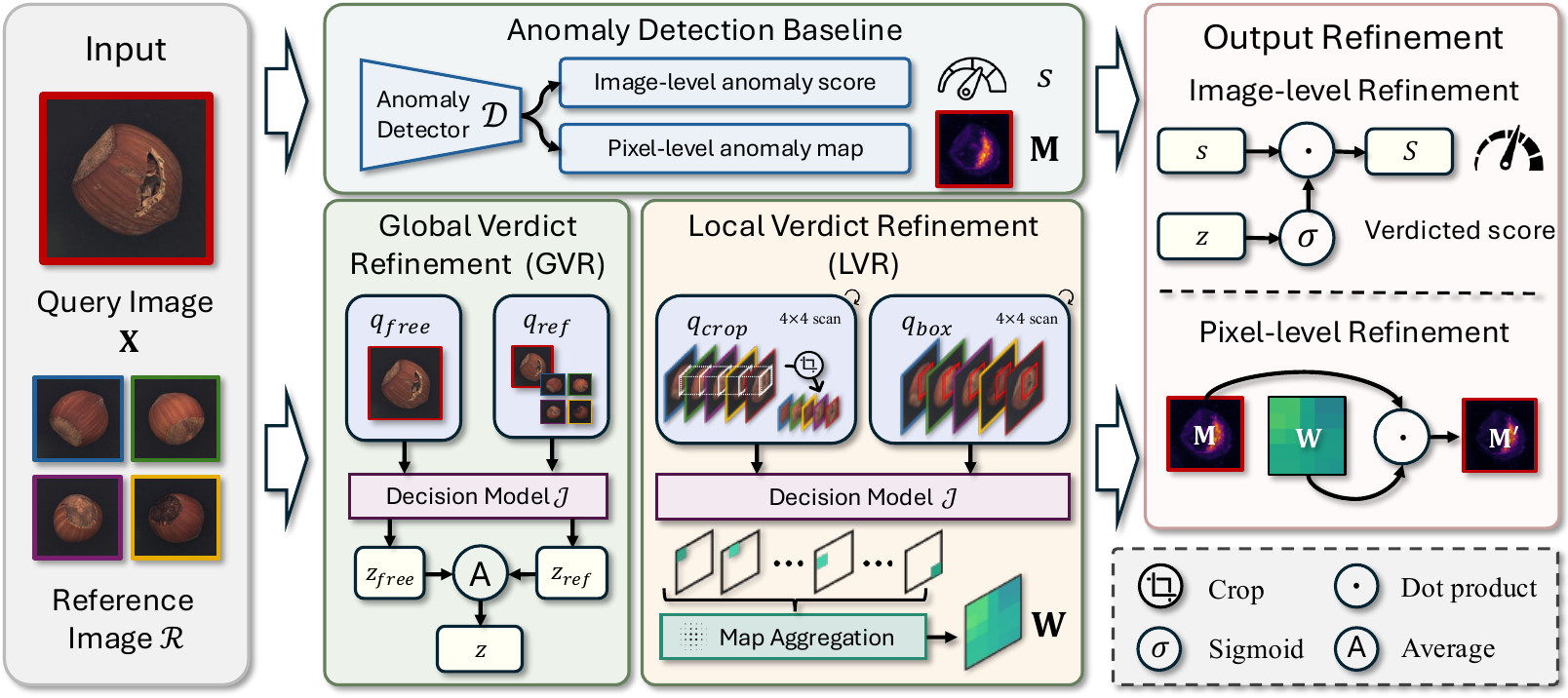}
  \vspace{-6mm}
  \caption{Overview of VerdictAD. A base detector $\mathcal{D}$ produces an anomaly map $\mathbf{M}$ and a score $s$ from the query $\mathbf{X}$ and the references $\mathcal{R}$. The frozen decision model $\mathcal{J}$ answers a reference-free and a reference-conditioned question on the whole image, giving the image logit $z$ that gates $s$ into $S$. It also answers one question per window $b_i$, either on a crop pair or on the full image with a red box, giving window logits $z_i$ that form the verification map $\mathbf{W}$ and reweight $\mathbf{M}$ into $\mathbf{M}'$.}
  \label{fig:pipeline}
  \vspace{-2mm}
\end{figure*}

%% file: sec/fig_sec/qualitative_comparison.tex
\begin{figure*}[t]
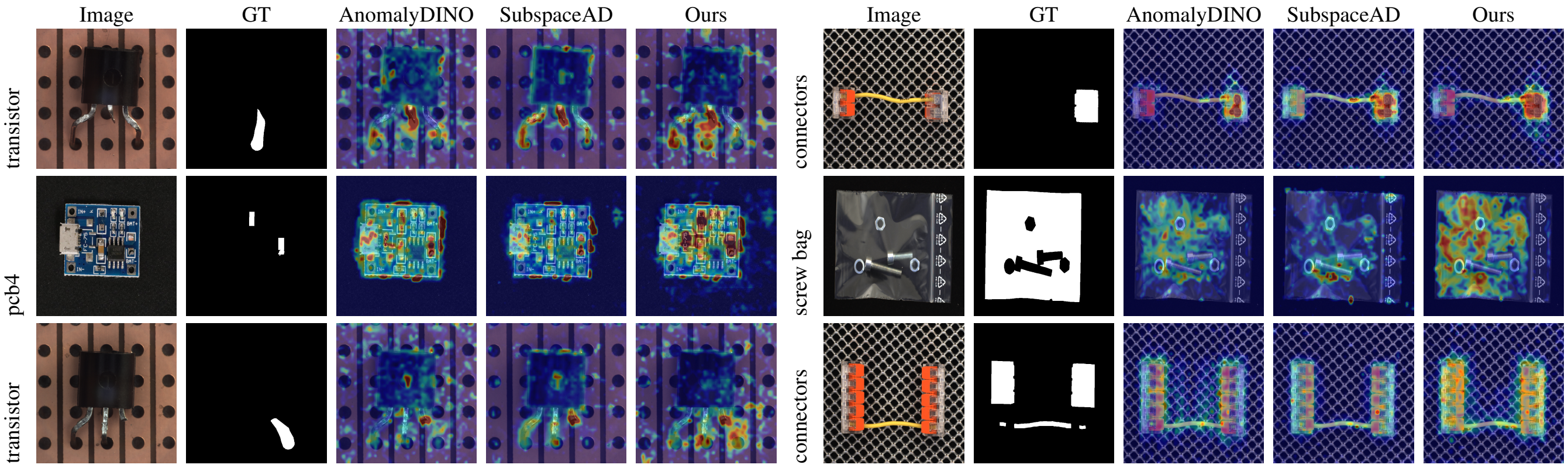

    \centering
    \setlength{\tabcolsep}{1.6pt}
    \renewcommand{\arraystretch}{1.0}

    \newcommand{\qroot}{sec/fig_sec/result_fig}
    \newcommand{\qimg}[2]{%
        \includegraphics[width=0.185\linewidth]{\qroot/#1/#2.jpg}%
    }
    \newcommand{\rowlabel}[1]{%
        \rotatebox{90}{\scriptsize\strut #1}%
    }

    \begin{minipage}[t]{0.49\textwidth}
        \centering
        \scriptsize
        \begin{tabular}{@{}cccccc@{}}
            & {Image} & {GT} & {AnomalyDINO} & {SubspaceAD} & {Ours} \\

            \rowlabel{transistor} &
            \qimg{showcase_mvtec_02_transistor_test_cut_lead_009}{input} &
            \qimg{showcase_mvtec_02_transistor_test_cut_lead_009}{gt} &
            \qimg{showcase_mvtec_02_transistor_test_cut_lead_009}{AnomalyDINO} &
            \qimg{showcase_mvtec_02_transistor_test_cut_lead_009}{SubspaceAD} &
            \qimg{showcase_mvtec_02_transistor_test_cut_lead_009}{Ours} \\
            
            \rowlabel{pcb4} &
            \qimg{showcase_visa_03_pcb4_test_bad_054}{input} &
            \qimg{showcase_visa_03_pcb4_test_bad_054}{gt} &
            \qimg{showcase_visa_03_pcb4_test_bad_054}{AnomalyDINO} &
            \qimg{showcase_visa_03_pcb4_test_bad_054}{SubspaceAD} &
            \qimg{showcase_visa_03_pcb4_test_bad_054}{Ours} \\

            \rowlabel{transistor} &
            \qimg{showcase_mvtec_04_transistor_test_cut_lead_001}{input} &
            \qimg{showcase_mvtec_04_transistor_test_cut_lead_001}{gt} &
            \qimg{showcase_mvtec_04_transistor_test_cut_lead_001}{AnomalyDINO} &
            \qimg{showcase_mvtec_04_transistor_test_cut_lead_001}{SubspaceAD} &
            \qimg{showcase_mvtec_04_transistor_test_cut_lead_001}{Ours} \\

        \end{tabular}

        \vspace{1mm}
        \textbf{(a) VisA and MVTec-AD}
    \end{minipage}
    \hfill
    \begin{minipage}[t]{0.49\textwidth}
        \centering
        \scriptsize
        \begin{tabular}{@{}cccccc@{}}
            & {Image} & {GT} & {AnomalyDINO} & {SubspaceAD} & {Ours} \\

            \rowlabel{connectors} &
            \qimg{showcase_loco_09_splicing_connectors_test_bad_splicing_connectors_structural_anomalies_063}{input} &
            \qimg{showcase_loco_09_splicing_connectors_test_bad_splicing_connectors_structural_anomalies_063}{gt} &
            \qimg{showcase_loco_09_splicing_connectors_test_bad_splicing_connectors_structural_anomalies_063}{AnomalyDINO} &
            \qimg{showcase_loco_09_splicing_connectors_test_bad_splicing_connectors_structural_anomalies_063}{SubspaceAD} &
            \qimg{showcase_loco_09_splicing_connectors_test_bad_splicing_connectors_structural_anomalies_063}{Ours} \\

            \rowlabel{screw bag} &
            \qimg{showcase_loco_07_screw_bag_test_bad_screw_bag_logical_anomalies_066}{input} &
            \qimg{showcase_loco_07_screw_bag_test_bad_screw_bag_logical_anomalies_066}{gt} &
            \qimg{showcase_loco_07_screw_bag_test_bad_screw_bag_logical_anomalies_066}{AnomalyDINO} &
            \qimg{showcase_loco_07_screw_bag_test_bad_screw_bag_logical_anomalies_066}{SubspaceAD} &
            \qimg{showcase_loco_07_screw_bag_test_bad_screw_bag_logical_anomalies_066}{Ours} \\

            \rowlabel{connectors} &
            \qimg{showcase_loco_03_splicing_connectors_test_bad_splicing_connectors_logical_anomalies_082}{input} &
            \qimg{showcase_loco_03_splicing_connectors_test_bad_splicing_connectors_logical_anomalies_082}{gt} &
            \qimg{showcase_loco_03_splicing_connectors_test_bad_splicing_connectors_logical_anomalies_082}{AnomalyDINO} &
            \qimg{showcase_loco_03_splicing_connectors_test_bad_splicing_connectors_logical_anomalies_082}{SubspaceAD} &
            \qimg{showcase_loco_03_splicing_connectors_test_bad_splicing_connectors_logical_anomalies_082}{Ours} \\
        \end{tabular}

        \vspace{1mm}
        \textbf{(b) MVTec LOCO AD}
    \end{minipage}

    \vspace{-1.5mm}
    \caption{
    Qualitative comparison of anomaly localization results on VisA, MVTec-AD, and MVTec LOCO AD.
    Each example shows the input image, ground truth (GT), and anomaly-map overlays from AnomalyDINO, SubspaceAD, and \method.
    Compared with the baseline detectors, \method generally produces more focused responses and suppresses distracting activations on irrelevant regions, while preserving the responses around true defects.
    The LOCO examples further illustrate that \method can improve localization quality for both structural and logical anomalies when the original anomaly map already contains useful evidence.
    }
    \label{fig:qualitative_comparison}
\end{figure*}

%% file: sec/4_experiment.tex
\input{sec/tab_sec/main_fewshot}

\section{Experiments}

\subsection{Experimental Settings}
\label{sec:impl}

\noindent\textbf{Datasets.}
We evaluate \method on three benchmarks: MVTec AD~\cite{bergmann2019mvtec}, VisA~\cite{zou2022spot}, and MVTec LOCO AD~\cite{bergmann2022beyond}.
MVTec AD contains 15 object and texture categories, while VisA consists of 12 industrial object categories.
MVTec LOCO AD contains five categories with both structural and logical anomalies, making it suitable for evaluating defects beyond local appearance differences.
All three datasets provide normal training images and test images with image-level and pixel-level anomaly annotations.
Following the few-shot protocol, we randomly select $k\in\{1,2,4\}$ normal images per category as references.

\vspace{1mm}
\noindent\textbf{Evaluation Metrics.}
Image-level performance is measured by AUROC and Average Precision (AUPR), while pixel-level performance is evaluated using pixel-wise AUROC and Area Under the Per-Region Overlap curve (AUPRO).
All metrics are reported as percentages.
For MVTec LOCO AD, we report standard AUPRO rather than the dataset-specific sPRO metric.

\vspace{1mm}
\noindent\textbf{Implementation Details.}
We use  9B-parameter Clef-Flash as the frozen decision model $\mathcal{J}$.
Our default detector extracts patch features from the 18th transformer layer of DINOv2-L.
All decision questions are binary, with positive answers indicating a defect, contamination, or incorrect arrangement.
Image-level inputs are resized to $896\times896$, while regional inputs are resized to $448\times448$.
For GVR, we set $\alpha=0.5$ for image-level refinement.
For LVR, we use $N=16$ overlapping windows on a $4\times4$ grid, with a normalized side length of $0.4$ and stride of $0.2$.
Both cropped and boxed views are used, with $\gamma=0.25$ for pixel-level refinement.
We use the same normal references as the corresponding base detector.
For reference-conditioned decisions, the references are arranged into a $2\times2$ mosaic, with cyclic repetition when $k<4$.
Each few-shot setting is evaluated using 3 random seeds for reference selection.

\vspace{1mm}
\noindent\textbf{Baselines.}
We compare \method with representative few-shot AD methods, including SPADE~\cite{cohen2020sub}, PatchCore~\cite{roth2022towards}, WinCLIP~\cite{jeong2023winclip}, PromptAD~\cite{li2024promptad}, IIPAD~\cite{lv2025one}, AnomalyDINO~\cite{damm2025anomalydino}, and SubspaceAD~\cite{lendering2026subspacead}.
To evaluate plug-and-play generality, we apply GVR and LVR to PatchCore, AnomalyDINO, and SubspaceAD without modifying their architectures.
For zero-shot evaluation, we consider AnomalyCLIP~\cite{zhou2024anomalyclip} and AdaCLIP~\cite{cao2024adaclip}.

%% file: sec/tab_sec/main_fewshot.tex
\newcommand{\MainResult}[3]{%
  \makebox[2.5em][l]{$#1#2$}%
  \makebox[2.75em][l]{\if\relax\detokenize{#3}\relax\else$\!_{\scriptscriptstyle\pm #3}$\fi}%
}
\begin{table*}[t]
\centering
\caption{Few-shot anomaly detection and localization on MVTec AD and VisA (\%). 
Values with subscripts report mean and standard deviation. 
Bold and underlined means indicate the best and second-best results within each shot and metric; the final four rows in each shot are excluded from ranking. 
$\Delta$ denotes the difference from the corresponding baseline in percentage points.}
\label{tab:main_fewshot}
\vspace{-2.mm}
\setlength{\tabcolsep}{2.5pt}
\renewcommand{\arraystretch}{1.10}
\footnotesize
\resizebox{\textwidth}{!}{%
\begin{tabular}{llcccccccc}
\toprule
& & \multicolumn{4}{c}{\textbf{MVTec AD}} & \multicolumn{4}{c}{\textbf{VisA}} \\
\cmidrule(lr){3-6}\cmidrule(lr){7-10}
& & \multicolumn{2}{c}{Image-level} & \multicolumn{2}{c}{Pixel-level} & \multicolumn{2}{c}{Image-level} & \multicolumn{2}{c}{Pixel-level} \\
\cmidrule(lr){3-4}\cmidrule(lr){5-6}\cmidrule(lr){7-8}\cmidrule(lr){9-10}
\textbf{Setup} & \textbf{Method}
& \multicolumn{1}{l}{AUROC}
& \multicolumn{1}{l}{AUPR}
& \multicolumn{1}{l}{AUROC}
& \multicolumn{1}{l}{AUPRO}
& \multicolumn{1}{l}{AUROC}
& \multicolumn{1}{l}{AUPR}
& \multicolumn{1}{l}{AUROC}
& \multicolumn{1}{l}{AUPRO} \\
\midrule
\multirow{13}{*}{1-shot} & SPADE~\cite{cohen2020sub} & \MainResult{81.0}{}{} & \MainResult{90.6}{}{} & \MainResult{91.2}{}{} & \MainResult{83.9}{}{} & \MainResult{79.5}{}{} & \MainResult{82.0}{}{} & \MainResult{95.6}{}{} & \MainResult{84.1}{}{} \\
 & PatchCore~\cite{roth2022towards} & \MainResult{83.4}{}{} & \MainResult{92.2}{}{} & \MainResult{92.0}{}{} & \MainResult{79.7}{}{} & \MainResult{79.9}{}{} & \MainResult{82.8}{}{} & \MainResult{95.4}{}{} & \MainResult{80.5}{}{} \\
 & WinCLIP~\cite{jeong2023winclip} & \MainResult{93.1}{}{} & \MainResult{96.5}{}{} & \MainResult{95.2}{}{} & \MainResult{87.1}{}{} & \MainResult{83.8}{}{} & \MainResult{85.1}{}{} & \MainResult{96.4}{}{} & \MainResult{85.1}{}{} \\
 & PromptAD~\cite{li2024promptad} & \MainResult{94.6}{}{} & \MainResult{97.1}{}{} & \MainResult{95.9}{}{} & \MainResult{87.9}{}{} & \MainResult{86.9}{}{} & \MainResult{88.4}{}{} & \MainResult{96.7}{}{} & \MainResult{85.1}{}{} \\
 & IIPAD~\cite{lv2025one} & \MainResult{94.2}{}{} & \MainResult{97.2}{}{} & \MainResult{96.4}{}{} & \MainResult{89.8}{}{} & \MainResult{85.4}{}{} & \MainResult{87.5}{}{} & \MainResult{96.9}{}{} & \MainResult{87.3}{}{} \\
 & DINOv2-L~\cite{oquab2023dinov2} & \MainResult{94.3}{}{0.3} & \MainResult{97.1}{}{0.1} & \MainResult{96.6}{}{0.0} & \MainResult{93.5}{}{0.2} & \MainResult{92.0}{}{0.4} & \MainResult{92.1}{}{0.3} & \MainResult{97.7}{}{0.1} & \MainResult{\underline{93.0}}{}{0.2} \\
 & AnomalyDINO~\cite{damm2025anomalydino} & \MainResult{96.6}{}{} & \MainResult{98.2}{}{} & \MainResult{96.8}{}{} & \MainResult{92.7}{}{} & \MainResult{87.4}{}{} & \MainResult{89.0}{}{} & \MainResult{97.8}{}{} & \MainResult{92.5}{}{} \\
 & SubspaceAD~\cite{lendering2026subspacead} & \MainResult{\underline{97.1}}{}{0.9} & \MainResult{\underline{98.6}}{}{0.4} & \MainResult{\mathbf{97.5}}{}{0.1} & \MainResult{\underline{93.7}}{}{0.2} & \MainResult{\underline{93.4}}{}{0.7} & \MainResult{\underline{93.1}}{}{0.8} & \MainResult{\underline{98.2}}{}{0.1} & \MainResult{\mathbf{93.5}}{}{0.2} \\
 & \cellcolor{verdictblue}\textbf{DINOv2-L + Ours} & \cellcolor{verdictblue}\MainResult{\mathbf{97.5}}{}{0.2} & \cellcolor{verdictblue}\MainResult{\mathbf{99.0}}{}{0.1} & \cellcolor{verdictblue}\MainResult{\underline{97.4}}{}{0.0} & \cellcolor{verdictblue}\MainResult{\mathbf{94.7}}{}{0.1} & \cellcolor{verdictblue}\MainResult{\mathbf{95.7}}{}{0.4} & \cellcolor{verdictblue}\MainResult{\mathbf{96.3}}{}{0.3} & \cellcolor{verdictblue}\MainResult{\mathbf{98.3}}{}{0.1} & \cellcolor{verdictblue}\MainResult{\mathbf{93.5}}{}{0.3} \\
\cmidrule(lr){2-10}
 & \cellcolor{verdictblue}\textbf{AnomalyDINO + Ours} & \cellcolor{verdictblue}\MainResult{98.1}{}{0.2} & \cellcolor{verdictblue}\MainResult{99.3}{}{0.1} & \cellcolor{verdictblue}\MainResult{97.4}{}{0.1} & \cellcolor{verdictblue}\MainResult{93.6}{}{0.1} & \cellcolor{verdictblue}\MainResult{95.2}{}{0.1} & \cellcolor{verdictblue}\MainResult{95.9}{}{0.2} & \cellcolor{verdictblue}\MainResult{98.0}{}{0.0} & \cellcolor{verdictblue}\MainResult{94.1}{}{0.1} \\
 & \cellcolor{verdictblue}\quad $\Delta$ & \cellcolor{verdictblue}\MainResult{+1.5}{}{} & \cellcolor{verdictblue}\MainResult{+1.1}{}{} & \cellcolor{verdictblue}\MainResult{+0.6}{}{} & \cellcolor{verdictblue}\MainResult{+0.9}{}{} & \cellcolor{verdictblue}\MainResult{+7.8}{}{} & \cellcolor{verdictblue}\MainResult{+6.9}{}{} & \cellcolor{verdictblue}\MainResult{+0.2}{}{} & \cellcolor{verdictblue}\MainResult{+1.6}{}{} \\
\addlinespace[1pt]
 & \cellcolor{verdictblue}\textbf{SubspaceAD + Ours} & \cellcolor{verdictblue}\MainResult{98.0}{}{0.4} & \cellcolor{verdictblue}\MainResult{99.3}{}{0.2} & \cellcolor{verdictblue}\MainResult{97.9}{}{0.1} & \cellcolor{verdictblue}\MainResult{95.2}{}{0.2} & \cellcolor{verdictblue}\MainResult{95.9}{}{0.3} & \cellcolor{verdictblue}\MainResult{96.4}{}{0.3} & \cellcolor{verdictblue}\MainResult{98.7}{}{0.0} & \cellcolor{verdictblue}\MainResult{95.9}{}{0.1} \\
 & \cellcolor{verdictblue}\quad $\Delta$ & \cellcolor{verdictblue}\MainResult{+0.9}{}{} & \cellcolor{verdictblue}\MainResult{+0.7}{}{} & \cellcolor{verdictblue}\MainResult{+0.4}{}{} & \cellcolor{verdictblue}\MainResult{+1.5}{}{} & \cellcolor{verdictblue}\MainResult{+2.5}{}{} & \cellcolor{verdictblue}\MainResult{+3.3}{}{} & \cellcolor{verdictblue}\MainResult{+0.5}{}{} & \cellcolor{verdictblue}\MainResult{+2.4}{}{} \\
\midrule
\multirow{13}{*}{2-shot} & SPADE~\cite{cohen2020sub} & \MainResult{82.9}{}{} & \MainResult{91.7}{}{} & \MainResult{92.0}{}{} & \MainResult{85.7}{}{} & \MainResult{80.7}{}{} & \MainResult{82.3}{}{} & \MainResult{96.2}{}{} & \MainResult{85.7}{}{} \\
 & PatchCore~\cite{roth2022towards} & \MainResult{86.3}{}{} & \MainResult{93.8}{}{} & \MainResult{93.3}{}{} & \MainResult{82.3}{}{} & \MainResult{81.6}{}{} & \MainResult{84.8}{}{} & \MainResult{96.1}{}{} & \MainResult{82.6}{}{} \\
 & WinCLIP~\cite{jeong2023winclip} & \MainResult{94.4}{}{} & \MainResult{97.0}{}{} & \MainResult{96.0}{}{} & \MainResult{88.4}{}{} & \MainResult{84.6}{}{} & \MainResult{85.8}{}{} & \MainResult{96.8}{}{} & \MainResult{86.2}{}{} \\
 & PromptAD~\cite{li2024promptad} & \MainResult{95.7}{}{} & \MainResult{97.9}{}{} & \MainResult{96.2}{}{} & \MainResult{88.5}{}{} & \MainResult{88.3}{}{} & \MainResult{90.0}{}{} & \MainResult{97.1}{}{} & \MainResult{85.8}{}{} \\
 & IIPAD~\cite{lv2025one} & \MainResult{95.7}{}{} & \MainResult{97.9}{}{} & \MainResult{96.7}{}{} & \MainResult{90.3}{}{} & \MainResult{86.7}{}{} & \MainResult{88.6}{}{} & \MainResult{97.2}{}{} & \MainResult{87.9}{}{} \\
 & DINOv2-L~\cite{oquab2023dinov2} & \MainResult{95.4}{}{0.5} & \MainResult{97.5}{}{0.3} & \MainResult{97.0}{}{0.1} & \MainResult{\underline{94.3}}{}{0.1} & \MainResult{92.5}{}{0.8} & \MainResult{92.5}{}{0.7} & \MainResult{98.1}{}{0.0} & \MainResult{\underline{93.6}}{}{0.3} \\
 & AnomalyDINO~\cite{damm2025anomalydino} & \MainResult{96.9}{}{} & \MainResult{98.2}{}{} & \MainResult{97.0}{}{} & \MainResult{93.1}{}{} & \MainResult{89.7}{}{} & \MainResult{90.7}{}{} & \MainResult{98.0}{}{} & \MainResult{93.4}{}{} \\
 & SubspaceAD~\cite{lendering2026subspacead} & \MainResult{\underline{97.6}}{}{0.6} & \MainResult{\underline{98.8}}{}{0.4} & \MainResult{\mathbf{97.7}}{}{0.1} & \MainResult{94.1}{}{0.1} & \MainResult{\underline{93.7}}{}{0.5} & \MainResult{\underline{93.3}}{}{0.8} & \MainResult{\underline{98.3}}{}{0.0} & \MainResult{\underline{93.6}}{}{0.1} \\
 & \cellcolor{verdictblue}\textbf{DINOv2-L + Ours} & \cellcolor{verdictblue}\MainResult{\mathbf{98.2}}{}{0.2} & \cellcolor{verdictblue}\MainResult{\mathbf{99.3}}{}{0.0} & \cellcolor{verdictblue}\MainResult{\underline{97.6}}{}{0.0} & \cellcolor{verdictblue}\MainResult{\mathbf{95.1}}{}{0.0} & \cellcolor{verdictblue}\MainResult{\mathbf{96.5}}{}{0.3} & \cellcolor{verdictblue}\MainResult{\mathbf{96.9}}{}{0.3} & \cellcolor{verdictblue}\MainResult{\mathbf{98.5}}{}{0.0} & \cellcolor{verdictblue}\MainResult{\mathbf{94.2}}{}{0.4} \\
\cmidrule(lr){2-10}
 & \cellcolor{verdictblue}\textbf{AnomalyDINO + Ours} & \cellcolor{verdictblue}\MainResult{98.6}{}{0.2} & \cellcolor{verdictblue}\MainResult{99.5}{}{0.0} & \cellcolor{verdictblue}\MainResult{97.6}{}{0.1} & \cellcolor{verdictblue}\MainResult{94.0}{}{0.2} & \cellcolor{verdictblue}\MainResult{95.9}{}{0.4} & \cellcolor{verdictblue}\MainResult{96.4}{}{0.4} & \cellcolor{verdictblue}\MainResult{98.2}{}{0.0} & \cellcolor{verdictblue}\MainResult{94.8}{}{0.2} \\
 & \cellcolor{verdictblue}\quad $\Delta$ & \cellcolor{verdictblue}\MainResult{+1.7}{}{} & \cellcolor{verdictblue}\MainResult{+1.3}{}{} & \cellcolor{verdictblue}\MainResult{+0.6}{}{} & \cellcolor{verdictblue}\MainResult{+0.9}{}{} & \cellcolor{verdictblue}\MainResult{+6.2}{}{} & \cellcolor{verdictblue}\MainResult{+5.7}{}{} & \cellcolor{verdictblue}\MainResult{+0.2}{}{} & \cellcolor{verdictblue}\MainResult{+1.4}{}{} \\
\addlinespace[1pt]
 & \cellcolor{verdictblue}\textbf{SubspaceAD + Ours} & \cellcolor{verdictblue}\MainResult{98.6}{}{0.2} & \cellcolor{verdictblue}\MainResult{99.5}{}{0.1} & \cellcolor{verdictblue}\MainResult{98.0}{}{0.0} & \cellcolor{verdictblue}\MainResult{95.4}{}{0.1} & \cellcolor{verdictblue}\MainResult{96.4}{}{0.3} & \cellcolor{verdictblue}\MainResult{96.8}{}{0.3} & \cellcolor{verdictblue}\MainResult{98.7}{}{0.0} & \cellcolor{verdictblue}\MainResult{96.2}{}{0.0} \\
 & \cellcolor{verdictblue}\quad $\Delta$ & \cellcolor{verdictblue}\MainResult{+1.0}{}{} & \cellcolor{verdictblue}\MainResult{+0.7}{}{} & \cellcolor{verdictblue}\MainResult{+0.3}{}{} & \cellcolor{verdictblue}\MainResult{+1.3}{}{} & \cellcolor{verdictblue}\MainResult{+2.7}{}{} & \cellcolor{verdictblue}\MainResult{+3.5}{}{} & \cellcolor{verdictblue}\MainResult{+0.4}{}{} & \cellcolor{verdictblue}\MainResult{+2.6}{}{} \\
\midrule
\multirow{13}{*}{4-shot} & SPADE~\cite{cohen2020sub} & \MainResult{84.8}{}{} & \MainResult{92.5}{}{} & \MainResult{92.7}{}{} & \MainResult{87.0}{}{} & \MainResult{81.7}{}{} & \MainResult{83.4}{}{} & \MainResult{96.6}{}{} & \MainResult{87.3}{}{} \\
 & PatchCore~\cite{roth2022towards} & \MainResult{88.8}{}{} & \MainResult{94.5}{}{} & \MainResult{94.3}{}{} & \MainResult{84.3}{}{} & \MainResult{85.3}{}{} & \MainResult{87.5}{}{} & \MainResult{96.8}{}{} & \MainResult{84.9}{}{} \\
 & WinCLIP~\cite{jeong2023winclip} & \MainResult{95.2}{}{} & \MainResult{97.3}{}{} & \MainResult{96.2}{}{} & \MainResult{89.0}{}{} & \MainResult{87.3}{}{} & \MainResult{88.8}{}{} & \MainResult{97.2}{}{} & \MainResult{87.6}{}{} \\
 & PromptAD~\cite{li2024promptad} & \MainResult{96.6}{}{} & \MainResult{98.5}{}{} & \MainResult{96.5}{}{} & \MainResult{90.5}{}{} & \MainResult{89.1}{}{} & \MainResult{90.8}{}{} & \MainResult{97.4}{}{} & \MainResult{86.2}{}{} \\
 & IIPAD~\cite{lv2025one} & \MainResult{96.1}{}{} & \MainResult{98.1}{}{} & \MainResult{97.0}{}{} & \MainResult{91.2}{}{} & \MainResult{88.3}{}{} & \MainResult{89.6}{}{} & \MainResult{97.4}{}{} & \MainResult{88.3}{}{} \\
 & DINOv2-L~\cite{oquab2023dinov2} & \MainResult{97.2}{}{0.5} & \MainResult{98.4}{}{0.3} & \MainResult{97.4}{}{0.0} & \MainResult{\underline{95.2}}{}{0.1} & \MainResult{93.6}{}{0.5} & \MainResult{93.3}{}{0.6} & \MainResult{98.3}{}{0.0} & \MainResult{93.8}{}{0.3} \\
 & AnomalyDINO~\cite{damm2025anomalydino} & \MainResult{97.7}{}{} & \MainResult{98.7}{}{} & \MainResult{97.2}{}{} & \MainResult{93.4}{}{} & \MainResult{92.6}{}{} & \MainResult{92.9}{}{} & \MainResult{98.2}{}{} & \MainResult{\underline{94.1}}{}{} \\
 & SubspaceAD~\cite{lendering2026subspacead} & \MainResult{\underline{97.9}}{}{0.4} & \MainResult{\underline{99.0}}{}{0.2} & \MainResult{\underline{97.8}}{}{0.1} & \MainResult{94.3}{}{0.0} & \MainResult{\underline{94.7}}{}{0.2} & \MainResult{\underline{94.3}}{}{0.5} & \MainResult{\underline{98.4}}{}{0.0} & \MainResult{93.8}{}{0.2} \\
 & \cellcolor{verdictblue}\textbf{DINOv2-L + Ours} & \cellcolor{verdictblue}\MainResult{\mathbf{98.7}}{}{0.3} & \cellcolor{verdictblue}\MainResult{\mathbf{99.5}}{}{0.1} & \cellcolor{verdictblue}\MainResult{\mathbf{97.9}}{}{0.0} & \cellcolor{verdictblue}\MainResult{\mathbf{95.7}}{}{0.0} & \cellcolor{verdictblue}\MainResult{\mathbf{96.9}}{}{0.1} & \cellcolor{verdictblue}\MainResult{\mathbf{97.2}}{}{0.1} & \cellcolor{verdictblue}\MainResult{\mathbf{98.7}}{}{0.0} & \cellcolor{verdictblue}\MainResult{\mathbf{94.4}}{}{0.3} \\
\cmidrule(lr){2-10}
 & \cellcolor{verdictblue}\textbf{AnomalyDINO + Ours} & \cellcolor{verdictblue}\MainResult{98.9}{}{0.2} & \cellcolor{verdictblue}\MainResult{99.6}{}{0.1} & \cellcolor{verdictblue}\MainResult{97.7}{}{0.1} & \cellcolor{verdictblue}\MainResult{94.2}{}{0.1} & \cellcolor{verdictblue}\MainResult{96.3}{}{0.1} & \cellcolor{verdictblue}\MainResult{96.7}{}{0.1} & \cellcolor{verdictblue}\MainResult{98.3}{}{0.0} & \cellcolor{verdictblue}\MainResult{95.2}{}{0.2} \\
 & \cellcolor{verdictblue}\quad $\Delta$ & \cellcolor{verdictblue}\MainResult{+1.2}{}{} & \cellcolor{verdictblue}\MainResult{+0.9}{}{} & \cellcolor{verdictblue}\MainResult{+0.5}{}{} & \cellcolor{verdictblue}\MainResult{+0.8}{}{} & \cellcolor{verdictblue}\MainResult{+3.7}{}{} & \cellcolor{verdictblue}\MainResult{+3.8}{}{} & \cellcolor{verdictblue}\MainResult{+0.1}{}{} & \cellcolor{verdictblue}\MainResult{+1.1}{}{} \\
\addlinespace[1pt]
 & \cellcolor{verdictblue}\textbf{SubspaceAD + Ours} & \cellcolor{verdictblue}\MainResult{98.8}{}{0.1} & \cellcolor{verdictblue}\MainResult{99.6}{}{0.0} & \cellcolor{verdictblue}\MainResult{98.1}{}{0.0} & \cellcolor{verdictblue}\MainResult{95.6}{}{0.1} & \cellcolor{verdictblue}\MainResult{96.5}{}{0.1} & \cellcolor{verdictblue}\MainResult{96.8}{}{0.1} & \cellcolor{verdictblue}\MainResult{98.8}{}{0.0} & \cellcolor{verdictblue}\MainResult{96.3}{}{0.1} \\
 & \cellcolor{verdictblue}\quad $\Delta$ & \cellcolor{verdictblue}\MainResult{+0.9}{}{} & \cellcolor{verdictblue}\MainResult{+0.6}{}{} & \cellcolor{verdictblue}\MainResult{+0.3}{}{} & \cellcolor{verdictblue}\MainResult{+1.3}{}{} & \cellcolor{verdictblue}\MainResult{+1.8}{}{} & \cellcolor{verdictblue}\MainResult{+2.5}{}{} & \cellcolor{verdictblue}\MainResult{+0.4}{}{} & \cellcolor{verdictblue}\MainResult{+2.5}{}{} \\
\bottomrule
\end{tabular}}
\vspace{-3.mm}
\end{table*}

%% file: sec/5_result.tex
\subsection{Comparison with State-of-the-Art Methods}
\label{sec:results_sota}

\input{sec/tab_sec/loco}

\noindent\textbf{Few-shot Anomaly Detection.}
Tab.~\ref{tab:main_fewshot} compares \method\ with existing methods on MVTec AD and VisA under the 1-, 2-, and 4-shot settings.
Despite using a simple DINOv2-L detector, \method\ achieves competitive or state-of-the-art performance across different few-shot settings, particularly in image-level detection.
Using SubspaceAD as the base detector further improves localization, achieving consistently strong pixel-level results.

\vspace{1mm}
\noindent\textbf{Logical Anomaly Detection.}
Tab.~\ref{tab:loco} presents the 4-shot results on MVTec LOCO AD.
Our refinement substantially improves image-level detection and pixel-level localization, particularly when combined with SubspaceAD.
Different base detectors perform best on different metrics, suggesting that their underlying anomaly representations remain important even after refinement.

\vspace{1mm}
\noindent\textbf{Qualitative Comparison.}
Fig.~\ref{fig:qualitative_comparison} compares the anomaly maps produced by AnomalyDINO, SubspaceAD, and \method\ against the ground-truth annotations.
Compared with existing detectors, \method\ suppresses irrelevant responses and produces anomaly maps that better highlight the actual defective regions.

\subsection{Generalization Across Anomaly Detectors}
\label{sec:results_generalization}

\input{sec/tab_sec/zeroshot}
\input{sec/fig_sec/score_distributions}
\input{sec/fig_sec/roc_ranking}

\input{sec/fig_sec/hyperparameters}

\vspace{1mm}
\noindent\textbf{Plug-and-Play Refinement.}
We apply \method\ to PatchCore, DINOv2-L, AnomalyDINO, and SubspaceAD without modifying their architectures.
Across all evaluated datasets and few-shot settings, all four detectors consistently improve in both image-level AUROC and pixel-level AUROC.
These improvements are observed across different reference selections, rather than being limited to a single run.
The examples in Fig.~\ref{fig:intro_examples} further illustrate how the same refinement suppresses misleading responses caused by appearance variations.

\vspace{1mm}
\noindent\textbf{Image-Score Distributions and Ranking.}
Fig.~\ref{fig:score_distributions} and~\ref{fig:roc_ranking} examine how GVR changes the distribution of image-level anomaly scores.
The refined scores provide better separation between normal and anomalous samples, particularly on MVTec LOCO AD and VisA.
Pairwise analysis further shows that GVR corrects many initially misordered normal--anomalous pairs while introducing relatively few new errors.

\vspace{1mm}
\noindent\textbf{Zero-Shot Transfer.}
We further apply reference-free GVR to AnomalyCLIP and AdaCLIP without normal references or local refinement.
As shown in Tab.~\ref{tab:zeroshot}, both detectors achieve higher image-level AUROC across all three benchmarks, with particularly large improvements on VisA.
Since only image-level scores are refined, pixel-level predictions remain unchanged.

\subsection{Ablation Study}
\label{sec:results_ablation}

\input{sec/tab_sec/ablation_global}
\input{sec/tab_sec/ablation_local}

\noindent\textbf{Global Verdict Refinement.}
Tab.~\ref{tab:ablation_global} compares reference-free, reference-conditioned, and combined decisions using the DINOv2-L detector.
Averaging the two logits consistently outperforms either decision alone.
Combining these decisions with the feature-based detector further improves image-level detection, indicating that typed decisions provide complementary information rather than replacing visual anomaly evidence.

\vspace{1mm}
\noindent\textbf{Local Verdict Refinement.}
Tab.~\ref{tab:ablation_local} compares crop-based, box-based, and combined regional decisions using SubspaceAD.
Box-based decisions are particularly effective for logical anomalies, where identifying a defect often requires understanding the surrounding context.
Combining both views achieves the best overall AUPRO, while the improvements on structural anomalies are more modest.
The refined maps also show higher defect-to-background response ratios, indicating that background responses are suppressed more strongly than defect-related ones.

\vspace{1mm}
\noindent\textbf{Hyperparameter Sensitivity.}
Fig.~\ref{fig:hyperparameters} examines the image-gating exponent, window exponent, window type, and DINOv2-L feature layer.
The selected exponents are supported by synthetic validation experiments, while the remaining analyses examine sensitivity to different configurations.
Test-set analyses are used for sensitivity assessment rather than hyperparameter selection.

%% file: sec/tab_sec/loco.tex
\begin{table}[t]
\centering
\caption{Four-shot anomaly detection and localization on MVTec LOCO AD (\%). Results are reported as mean and standard deviation over three seeds. Best and second-best results are bold and underlined.}
\label{tab:loco}
\setlength{\tabcolsep}{3.4pt}
\renewcommand{\arraystretch}{1.1}
\footnotesize
\resizebox{\columnwidth}{!}{%
\begin{tabular}{lcccc}
\toprule
& \multicolumn{2}{c}{Image-level}
& \multicolumn{2}{c}{Pixel-level} \\
\cmidrule(lr){2-3}\cmidrule(l){4-5}
Method & AUROC & AUPR & AUROC & AUPRO \\
\midrule

PatchCore~\cite{roth2022towards}
& $67.4_{\scriptscriptstyle\pm3.3}$
& $77.6_{\scriptscriptstyle\pm2.5}$
& $74.9_{\scriptscriptstyle\pm0.0}$
& $66.7_{\scriptscriptstyle\pm0.5}$ \\
\rowcolor{verdictblue}
PatchCore + Ours
& ${78.0}_{\scriptscriptstyle\pm2.4}$
& ${86.5}_{\scriptscriptstyle\pm1.5}$
& $79.9_{\scriptscriptstyle\pm0.3}$
& $68.7_{\scriptscriptstyle\pm0.4}$ \\

\midrule

DINOv2-L~\cite{oquab2023dinov2}
& $68.7_{\scriptscriptstyle\pm0.8}$
& $80.3_{\scriptscriptstyle\pm0.8}$
& $77.1_{\scriptscriptstyle\pm0.1}$
& $62.8_{\scriptscriptstyle\pm0.3}$ \\
\rowcolor{verdictblue}
DINOv2-L + Ours
& ${77.5}_{\scriptscriptstyle\pm0.1}$
& ${86.3}_{\scriptscriptstyle\pm0.1}$
& $80.0_{\scriptscriptstyle\pm0.1}$
& $65.4_{\scriptscriptstyle\pm0.2}$ \\

\midrule

AnomalyDINO~\cite{damm2025anomalydino}
& $67.4_{\scriptscriptstyle\pm2.0}$
& $77.7_{\scriptscriptstyle\pm1.2}$
& ${81.5}_{\scriptscriptstyle\pm0.2}$
& $59.5_{\scriptscriptstyle\pm0.8}$ \\
\rowcolor{verdictblue}
AnomalyDINO + Ours
& $74.9_{\scriptscriptstyle\pm1.7}$
& $83.3_{\scriptscriptstyle\pm0.7}$
& ${82.7}_{\scriptscriptstyle\pm0.3}$
& $61.1_{\scriptscriptstyle\pm1.0}$ \\

\midrule

SubspaceAD~\cite{lendering2026subspacead}
& $67.5_{\scriptscriptstyle\pm1.7}$
& $79.5_{\scriptscriptstyle\pm1.2}$
& $76.6_{\scriptscriptstyle\pm0.3}$
& ${69.5}_{\scriptscriptstyle\pm0.2}$ \\
\rowcolor{verdictblue}
SubspaceAD + Ours
& $75.3_{\scriptscriptstyle\pm2.0}$
& $84.7_{\scriptscriptstyle\pm1.2}$
& $79.3_{\scriptscriptstyle\pm0.1}$
& ${71.3}_{\scriptscriptstyle\pm0.6}$ \\

\bottomrule
\end{tabular}}
\end{table}

%% file: sec/tab_sec/zeroshot.tex
\begin{table}[t]
\centering
\caption{Zero-shot image-level AUROC and AUPR (\%) on three datasets.
Ours uses reference-free global decisions with a gating exponent of $0.25$.
Best and second-best results are bold and underlined.}
\label{tab:zeroshot}
\setlength{\tabcolsep}{3.3pt}
\renewcommand{\arraystretch}{1.1}
\footnotesize
\resizebox{\columnwidth}{!}{%
\begin{tabular}{lcccccc}
\toprule
& \multicolumn{2}{c}{MVTec AD} & \multicolumn{2}{c}{VisA} & \multicolumn{2}{c}{LOCO AD} \\
\cmidrule(lr){2-3}\cmidrule(lr){4-5}\cmidrule(l){6-7}
Method & AUROC & AUPR & AUROC & AUPR & AUROC & AUPR \\
\midrule
AnomalyCLIP & 91.55 & 96.36 & 81.97 & 85.34 & 61.02 & 74.75 \\
\rowcolor{verdictblue}
AnomalyCLIP + Ours & {92.97} & {97.19} & {90.70} & {92.82} & {65.10} & {79.14} \\
\addlinespace[1pt]
AdaCLIP & 90.08 & 95.74 & 86.31 & 88.68 & 60.38 & 73.39 \\
\rowcolor{verdictblue}
AdaCLIP + Ours & {92.39} & {97.09} & {92.08} & {93.71} & {64.80} & {78.51} \\
\bottomrule
\end{tabular}}
\end{table}

%% file: sec/fig_sec/score_distributions.tex
\definecolor{normalblue}{HTML}{2F6DB5}
\definecolor{anomalyorange}{HTML}{D9822B}

\begin{figure}[t]
\centering
\setlength{\tabcolsep}{1.5pt}
\renewcommand{\arraystretch}{1.0}

\begin{tabular}{@{}ccc@{}}
\multicolumn{3}{c}{
  \footnotesize
  \textcolor{normalblue}{\rule{1.2ex}{1.2ex}}\,
  Normal
  \qquad
  \textcolor{anomalyorange}{\rule{1.2ex}{1.2ex}}\,
  Anomalous
}\\
\includegraphics[width=.325\linewidth]{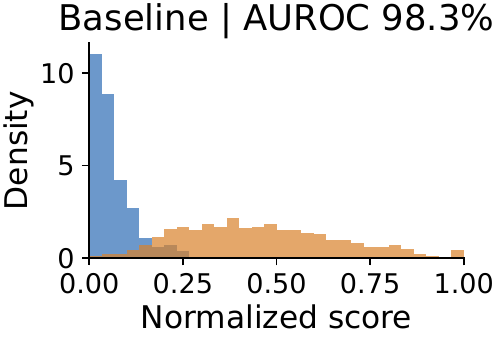} &
\includegraphics[width=.325\linewidth]{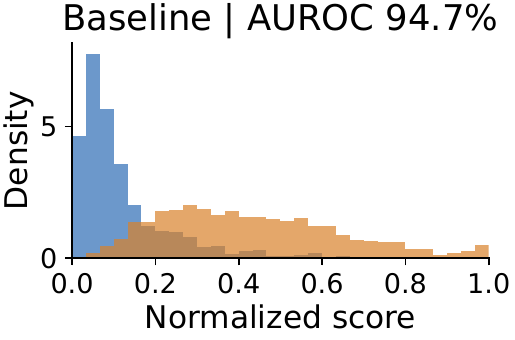} &
\includegraphics[width=.325\linewidth]{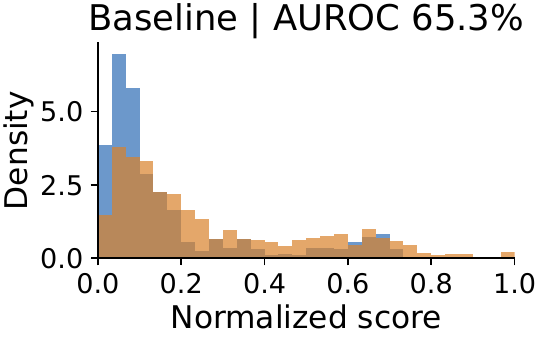} \\[-2pt]

\includegraphics[width=.325\linewidth]{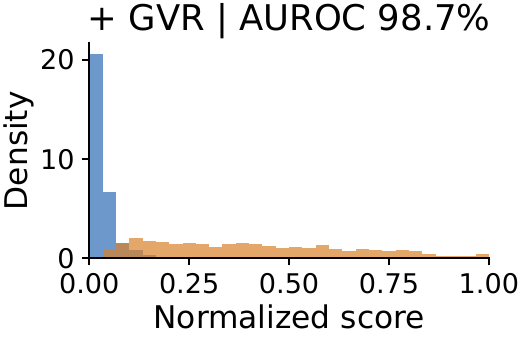} &
\includegraphics[width=.325\linewidth]{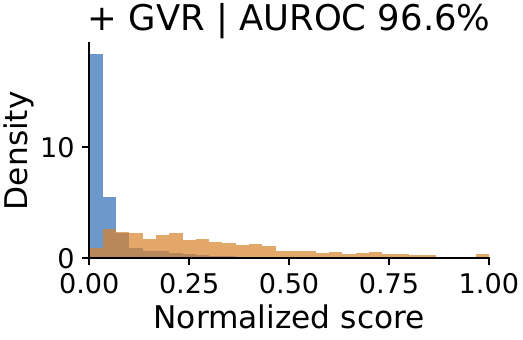} &
\includegraphics[width=.325\linewidth]{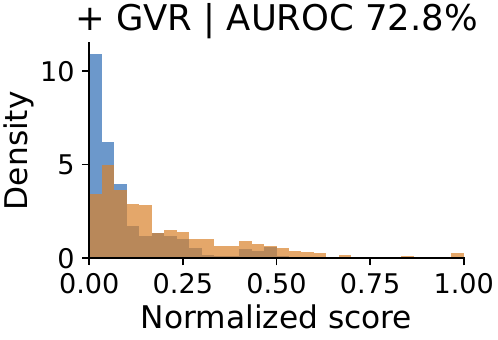} \\[-1pt]

{\scriptsize\textbf{(a) MVTec AD}} &
{\scriptsize\textbf{(b) VisA}} &
{\scriptsize\textbf{(c) MVTec LOCO AD}} \\[3pt]
\end{tabular}
\vspace{-3mm}
\caption{Image-level anomaly-score distributions
of SubspaceAD before (top) and after (bottom)
GVR under the 4-shot setting.
Scores are normalized within each class for
visualization.}
\label{fig:score_distributions}
\end{figure}

%% file: sec/fig_sec/roc_ranking.tex
\begin{figure*}[t]
\centering
\setlength{\tabcolsep}{3pt}
\begin{tabular}{@{}ccc@{}}
\includegraphics[width=.322\linewidth]{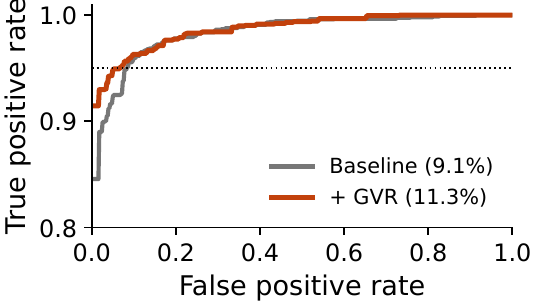} &
\includegraphics[width=.322\linewidth]{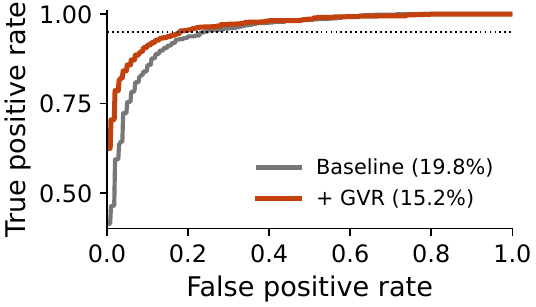} &
\includegraphics[width=.322\linewidth]{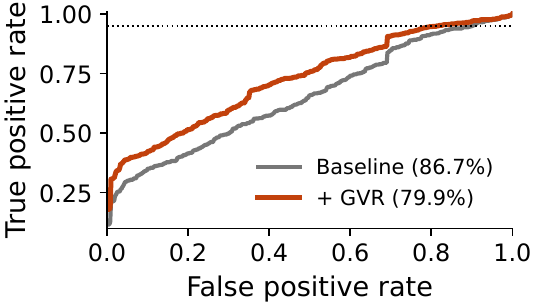} \\
{(a) MVTec AD} & {(b) VisA} & {(c) LOCO AD}
\end{tabular}
\vspace{-3pt}
\caption{Image-level ROC curves for SubspaceAD before and after GVR on three industrial anomaly detection benchmarks (4-shot, single-seed analysis). The ranking-correction statistics are discussed in the text; no separate ranking plot is included here.}
\label{fig:roc_ranking}
\end{figure*}

%% file: sec/fig_sec/hyperparameters.tex
\begin{figure*}[t]
\centering
\setlength{\tabcolsep}{4pt}
\begin{tabular}{@{}ccc@{}}
\includegraphics[width=.32\textwidth]{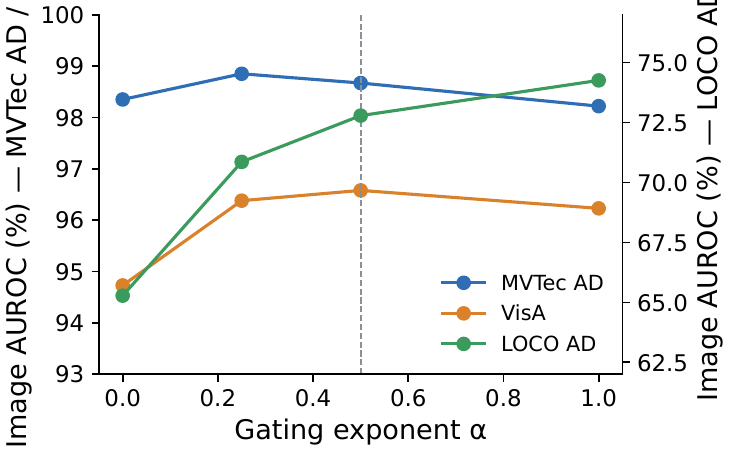} &
\includegraphics[width=.32\textwidth]{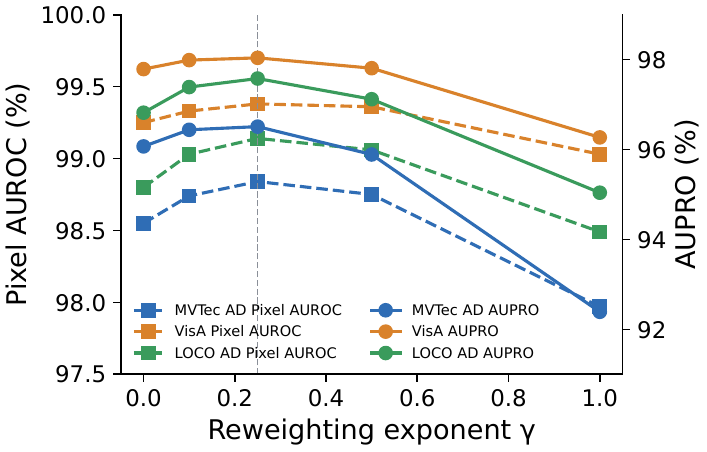} &
\includegraphics[width=.32\textwidth]{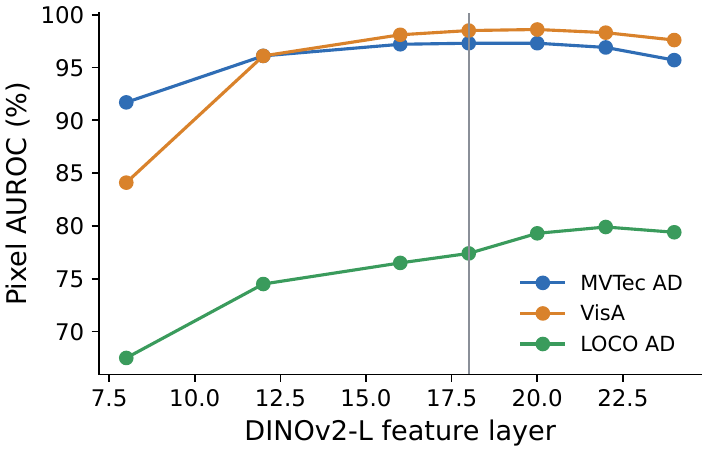} \\[-1pt]
\multicolumn{1}{c}{\small (a) Global gating exponent} &
\multicolumn{1}{c}{\small (b) Local reweighting exponent} &
\multicolumn{1}{c}{\small (c) DINOv2-L feature layer} \\
\end{tabular}
\vspace{-3pt}
\caption{Sensitivity analyses of \method.
(a) Effect of the global gating exponent $\alpha$, with separate vertical scales for MVTec AD/VisA and LOCO AD.
(b) Effect of the local reweighting exponent $\gamma$ on Pixel AUROC and AUPRO, evaluated on an independent synthetic validation set.
(c) Effect of the DINOv2-L feature layer on pixel-level detection performance.}
\label{fig:hyperparameters}
\end{figure*}

%% file: sec/tab_sec/ablation_global.tex
\begin{table}[t]
\centering
\caption{Ablation of global decisions. Image AUROC (\%).}
\label{tab:ablation_global}
\vspace{-3mm}
\setlength{\tabcolsep}{3pt}
\footnotesize
\resizebox{\columnwidth}{!}{%
\begin{tabular}{lccc}
\toprule
Configuration & MVTec AD & VisA & LOCO AD \\
\midrule
Reference-free decision & 91.8 & 89.8 & 63.3 \\
Reference-conditioned decision & 93.9 & 86.3 & 69.5 \\
Combined global decisions & 96.3 & 91.7 & 73.8 \\
DINOv2-L detector & 95.5 & 93.7 & 67.1 \\
\rowcolor{verdictblue}
DINOv2-L + GVR & \textbf{98.3} & \textbf{96.9} & \textbf{76.2} \\
\bottomrule
\end{tabular}}
\end{table}

%% file: sec/tab_sec/ablation_local.tex
\begin{table}[t]
\centering
\caption{Window-view ablation on MVTec LOCO AD, with the 4-shot SubspaceAD detector (single seed). Metrics are pixel AUROC / AUPRO (\%).}
\label{tab:ablation_local}
\setlength{\tabcolsep}{3pt}
\footnotesize
\resizebox{\columnwidth}{!}{%
\begin{tabular}{lccc}
\toprule
Windows & Logical & Structural & All \\
\midrule
No LVR & 75.98 / 60.77 & 94.22 / 89.20 & 77.05 / 69.34 \\
Crop only & 77.77 / 63.51 & 95.06 / 90.76 & 78.31 / 70.60 \\
Box only & 79.90 / 64.43 & 94.98 / 90.23 & 80.48 / 71.46 \\
\rowcolor{verdictblue}
Crop + Box & 79.31 / {64.70} & {95.21} / 90.67 & 79.87 / {71.60} \\
\bottomrule
\end{tabular}}
\end{table}

%% file: sec/6_conclu.tex
\section{Conclusion}

In this paper, we presented \method, a training-free framework that uses typed visual decisions to refine the outputs of existing anomaly detectors. Rather than learning new anomaly representations, \method focuses on verifying whether the detected visual differences correspond to genuine defects. We introduced Global Verdict Refinement (GVR) to improve image-level anomaly discrimination through reference-free and reference-conditioned decisions, and Local Verdict Refinement (LVR) to suppress misleading pixel-level responses using regional verification. Experiments on MVTec AD, VisA, and MVTec LOCO AD demonstrate strong performance across different few-shot settings and consistent improvements over multiple detection backbones, particularly for logical anomalies. While our approach cannot recover anomaly evidence entirely absent from the original detector, it provides a simple and effective way to improve existing anomaly predictions without additional training.

%% file: sec/X_LLM_usage.tex
\section*{LLM Usage Statement}
ChatGPT assisted with manuscript editing, language refinement,
LaTeX formatting, and table and visualization preparation.
The authors verified all research decisions, results, and
scientific claims and take full responsibility for the paper.